\documentclass[letterpaper]{article}

\PassOptionsToPackage{numbers, compress}{natbib}

     \usepackage[preprint]{neurips_2024}

\usepackage[utf8]{inputenc} 
\usepackage[T1]{fontenc}    
\usepackage{hyperref}       
\usepackage{url}
\usepackage[hyphenbreaks]{breakurl} 
\usepackage{booktabs}       
\usepackage{amsfonts}       
\usepackage{nicefrac}       
\usepackage{microtype}      
\usepackage{xcolor}         

\usepackage{amsmath}
\usepackage{amssymb}
\usepackage{mathtools}
\usepackage{amsthm}

\usepackage{hyperref}
\usepackage{threeparttable}
\usepackage{multirow}
\usepackage{pifont}
\usepackage{float}

\newcommand{\cmark}{\textcolor{green}{\ding{52}}}
\newcommand{\xmark}{\textcolor{red}{\ding{55}}}

\title{An Empirical Study of $\mu$P Learning Rate Transfer} 

\author{
  Lucas Lingle \\
  Independent Researcher \\
  \texttt{lucasdaxlingle@gmail.com} \\
}

\begin{document}

\maketitle

\begin{abstract}
Deep learning models have become a cornerstone of modern AI research, yet their initializations and learning rates may at times be set in an opaque or ad-hoc fashion due to the high cost of hyperparameter sweeps. 
The $\mu$-Parameterization ($\mu$P) offers a possible solution to this challenge, yielding scaling rules for model initialization and learning rates while reportedly enabling zero-shot hyperparameter transfer from small to large models. 
Despite its evident promise, the $\mu$P method is not yet widely adopted, perhaps due to higher implementation complexity, many variations, or complex theoretical background. 
This work considers $\mu$P empirically, focusing on the popular transformer architecture, and aims to answer a simple question: does $\mu$-Transfer yield near-optimal learning rates in practice? 
Studying over a dozen ablations with up to 1.2B parameters and 33B tokens and a large-scale experiment with up to 10B parameters and 190B tokens, we observe a positive answer for most settings, and discuss improvements otherwise. 
\end{abstract}

\section{Introduction}
In spite of transformer models emerging as a primary architecture for modern AI applications \citep{OpenAI2023GPT4, Anthropic2024Claude3, GeminiTeam2023Gemini1, Reid2024Gemini1dot5, Touvron2023Llama, Touvron2023Llama2, Jiang2024Mixtral, Parmar2024Nemotron4, Dehghani2023ViT22B, Liu2023VisionInstructionTuning, OpenAI2024GPT4o, Copet2023MusicGen, Dieleman2021VariableRate}, there is no standard method for setting their initialization, learning rate, or other hyperparameters. Further, optimizing the hyperparameters of large models is complicated by the expense of training multiple models of the target size, while naive selection may yield suboptimal results. 

The $\mu$-Parameterization ($\mu$P) \citep{Yang2021TP4, Yang2022TP5, Yang2023TP6, Yang2024SpectralCondition} offers one possible method for scaling initializations and learning rates, based on the Tensor Programs formalism of neural network training \citep{Yang2019TP1, Yang2020TP2, Yang2021TP3}. Empirically, $\mu$P is also reported to enable zero-shot hyperparameter transfer from small proxy models to large target models \citep{Yang2021TP4, Yang2022TP5, Yang2023TP6}, using width or depth as the direction of scaling. This `$\mu$-transfer' technique offers a promise of stable training and optimal hyperparameters at scale with low expense. 

However, the initial report on $\mu$-transfer demonstrated approximate preservation of hyperparameter optima mostly on a smaller scale, with the largest-scale experiment designed as a benchmark \citep{Yang2022TP5}. 
Thus, it remains unclear if hyperparameter optima are empirically preserved under $\mu$-transfer when target model is very large. For example, the learning rate optimum might drift or jump due to interaction between emergent outlier features and weight updates \citep{Yang2022TP5, Dettmers2022Emergent}. 

A second open question is if $\mu$-transfer is compatible with the techniques used in practice, such as decoupled weight decay \citep{Loshchilov2017AdamW} or multiplicative nonlinearities \citep{Shazeer2020GLU, So2021Primer}. While the initial report aims to describe the compatible techniques, there remains a need for further empirical study. 

Several recent works may have applied $\mu$P \citep{Dey2023CerebrasGPT, Dey2023BTLM3B8K, Liu2023LLM360, Hu2024MiniCPM, XAI2024Grok1}, but do not settle the open questions above; this would require extensive hyperparameter sweeps at scale. Inspired by the potential benefits, this paper aims to answer both the above questions. In particular, we perform a series of extensive ablations using models of up to 1.2B parameters and 33B tokens to answer the second question, as well as a larger-scale study using up to 10B parameter models and 190B tokens to answer the first. 

\section{Background and Notation}
This paper focuses on decoder-only transformer models \citep{Vaswani2017Transformer, Liu2018Decoder}, which process sequences of tokens $\mathbf{z} \in \{0, \ldots, V-1\}^{C}$, where $V$ is called the vocabulary size and $C$ the context length. This architecture has three components: the embedding layer, transformer layers, and unembedding layer. We describe a pre-norm transformer decoder \citep{Baevski2019AdaptiveInputs, Radford2019GPT2, Child2019SparseTransformers, Xiong2022PreLN} of depth $L$. 

\subsection{Embedding Layer}
The token sequence $\mathbf{z} \in \{0, \ldots, V-1\}^{C}$ is used to index into an {embedding matrix} $\mathbf{W}^{E} \in \mathbb{R}^{V \times M}$, where $M$ is called the model width. The resulting real-valued vectors are written as rows to an activation matrix $\mathbf{X}^{0} \in \mathbb{R}^{C \times M}$ according to the formula $\mathbf{X}_{i}^{0} = \mathbf{W}_{z_{i}}^{E}$. 

\subsection{Transformer Layers}
A transformer layer consists of two residual blocks \citep{He2016Identity}, denoted $\text{MHA}$ and $\text{MLP}$, which are added to a `residual stream' according to the formula
\begin{align}\label{layer}
\mathbf{X}^{\ell} = \mathbf{X}^{\ell-1} + \text{MHA}(\mathbf{X}^{\ell-1}) + \text{MLP}(\mathbf{X}^{\ell-1} + \text{MHA}(\mathbf{X}^{\ell-1})). 
\end{align}

The MHA residual block performs multi-head self-attention, defined by a head width $D \in \mathbb{N}$ and number of heads $H \in \mathbb{N}$. For each head, MHA uses a distinct set of projections $\mathbf{W}^{AQ}, \mathbf{W}^{AK}, \mathbf{W}^{AV} \in \mathbb{R}^{M \times D}$ to perform the following computations given input $\mathbf{X} \in \mathbb{R}^{C \times M}$:
\begin{align}
\mathbf{Y} &= \text{LayerNorm}(\mathbf{X}) \\
\mathbf{Q} &= \mathbf{Y}\mathbf{W}^{AQ} \\
\mathbf{K} &= \mathbf{Y}\mathbf{W}^{AK} \\
\mathbf{V} &= \mathbf{Y}\mathbf{W}^{AV} \\
\mathbf{S} &= \tau^{-1} \mathbf{Q}\mathbf{K}^{\top} + \mathbf{M} \\
\mathbf{P} &= \text{Softmax}(\mathbf{S}) \\
\mathbf{O} &= \mathbf{P}\mathbf{V} 
\end{align}
where the LayerNorm and Softmax functions are applied row-wise, $\tau^{-1} > 0$ is a constant typically set to $1/\sqrt{D}$, and $\mathbf{M}$ is a causal mask given by $\mathbf{M}_{i, j} = -\infty$ if $j > i$ and $\mathbf{M}_{i, j} = 0$ otherwise. 
The heads' outputs $\mathbf{O}$ are concatenated together, and then projected using one additional matrix $\mathbf{W}^{AO} \in \mathbb{R}^{HD \times M}$ to form the residual $\text{MHA}(\mathbf{X})$. This residual is summed onto the residual stream as in Equation \ref{layer}, and the sum is processed by the MLP residual block. 

The MLP residual block applies a multi-layer perceptron to each sequence position independently. It is defined via hidden width $F$ and element-wise activation $\phi$. It uses two trainable projections $\mathbf{W}^{FI} \in \mathbb{R}^{M \times F}$ and $\mathbf{W}^{FO} \in \mathbb{R}^{F \times M}$. Given an input tensor $\mathbf{X}$, it defines the residual:
\begin{align}
\mathbf{Y} &= \text{LayerNorm}(\mathbf{X}) \\
\mathbf{O} &=  \phi(\mathbf{Y}\mathbf{W}^{FI})\mathbf{W}^{FO} 
\end{align}
This residual is likewise summed onto the residual stream, following Equation \ref{layer}. 

\subsection{Unembedding Layer}
The unembedding layer uses a matrix $\mathbf{W}^{U} \in \mathbb{R}^{M \times V}$ to produce the probabilities for next-token prediction. The layer's input is the residual stream output $\mathbf{X}^{L}$, and its output is
\begin{align}
\mathbf{Y} &= \text{LayerNorm}(\mathbf{X}^{L}) \\
\mathbf{U} &= \text{Softmax}(\mathbf{Y}\mathbf{W}^{U}) 
\end{align}
Due to the softmax, each row of $\mathbf{U} \in \mathbb{R}^{C \times V}$ defines a probability mass function over tokens in the vocabulary. The model is trained on the cross-entropy loss $-\frac{1}{C} \sum_{i=0}^{C-1} \log \mathbf{U}_{i, \mathbf{z}_{i+1}}$. 

\section{$\mu$-Transfer}
The $\mu$-Parameterization ($\mu$P) \citep{Yang2021TP4, Yang2022TP5, Yang2023TP6, Yang2024SpectralCondition} refers to a specific family of initializations and learning rates, empirically reported to facilitate hyperparameter transfer from small to large models (`$\mu$-Transfer'). This paper investigates $\mu$P for transformers with respect to width; we do not use depthwise $\mu$P \citep{Yang2023TP6} since it requires only one linear layer stacked per residual block, while transformers use two. 

The general formulation of $\mu$P when training with Adam \citep{Kingma2015Adam} and using an i.i.d. Gaussian initialization is given by \citet{Yang2022TP5}. The first three columns of Table \ref{rules-table} display these rules for transformers. 
These columns use big-theta notation. Formally, $f(x) = \Theta(g(x))$ if there exists $x_{0} \in \mathbb{R}$ and $c, C > 0$ s.t. $cg(x) \leq f(x) \leq Cg(x)$ for all $x > x_{0}$. 

\begin{table}[h]
\caption{$\mu$P scaling rules for transformers; a rule for attention scale is detailed in the main text.}
\label{rules-table}
\begin{center}
\begin{tabular}{c|cc|cc}
\toprule
\multicolumn{1}{c}{\bf Param}  &\multicolumn{1}{c}{\bf Init Variance ($\Theta$)} &\multicolumn{1}{c}{\bf Adam LR ($\Theta$)} &\multicolumn{1}{c}{\bf Init Variance (Exact)} &\multicolumn{1}{c}{\bf Adam LR (Exact)} \\
\midrule
$\mathbf{W}^{E}$ & $1$ & $1$ & $1$ & $\alpha$ \\
\midrule
$\mathbf{W}^{AQ}$ & $1/M$ & $1/M$ & $1/M$ & $\alpha P/M$ \\
$\mathbf{W}^{AK}$ & $1/M$ & $1/M$ & $1/M$ & $\alpha P/M$ \\
$\mathbf{W}^{AV}$ & $1/M$ & $1/M$ & $1/M$ & $\alpha P/M$ \\
$\mathbf{W}^{AO}$ & $1/(HD)$ & $1/(HD)$ & $1/M$ & $\alpha P/M$ \\
\midrule
$\mathbf{W}^{FI}$ & $1/M$ & $1/M$ & $1/M$ & $\alpha P/M$ \\
$\mathbf{W}^{FO}$ & $1/F$ & $1/F$ & $0.25/M$ & $\alpha P/M$ \\
\midrule
$\mathbf{W}^{U}$ & $1/M^{2}$ & $1/M$ & $1/M^{2}$ & $\alpha P/M$ \\
\bottomrule
\end{tabular}
\end{center}
\end{table}

In the remainder of this paper, we assume $HD = M$ and $F = 4M$ unless specifically mentioned. We fix a proxy model width $P = 128$ and head width $D = 128$,\footnote{We consider other choices of $D$ in Appendix \ref{extra-heads-appdx}.} and follow the specific scaling rules in the last two columns of Table \ref{rules-table}, where $\alpha$ denotes the \textbf{base learning rate}. The base learning rate is so named because it is the learning rate for all parameters when $M = P$. These relative scaling rules are a special case of those in Appendix B.1 of \citet{Yang2022TP5} and were chosen to make the initialization similar to a standard parameterization (only the unembedding initialization differs), and to make learning rates as uniform as possible (only the embedding learning rate differs). 

In addition, $\mu$P uses an attention scale of $\tau^{-1} = \Theta(1/D)$ instead of the usual $\tau^{-1} = 1/\sqrt{D}$. For simplicity, we use $\tau^{-1} = 1/D$, since in preliminary experiments we observed only a small improvement from using smaller multiples of $1/D$. Note that for $D$ fixed across model widths $M$, any constant $\tau^{-1} \neq 0$ technically complies with $\mu$P \citep{Yang2022TP5} but in the experiments, $\tau^{-1}$ will be shown to have a major impact on performance and transfer. 

It is also possible to add scalar multipliers throughout the network as hyperparameters, or to scale the initializations by constants. For simplicity, we focus on $\mu$-Transfer of the base learning rate. 

\section{Experiments}
\label{experiments}
\subsection{Experimental Setup}
\label{eval_proto}
\subsubsection{Implementation}
Our experiments are implemented using Jax/Flax \citep{Bradbury2018, Heek2023}. 
Training is performed on TPU V3 pod slices with 128-256 cores, using the fully-sharded data parallelism (FSDP) strategy from \citet{Xu2021GSPMD} to reduce memory overhead. 
Models train on the Colossal Clean Common Crawl (C4) dataset, using the T5 tokenizer \citep{Raffel2020T5} with context length $C = 256$.

The experiments use a bitwise-deterministic training pipeline, with shards of data written to disk in a random-access format similar to Nvidia Megatron \citep{Shoeybi2019Megatron}. Model checkpoints are saved periodically, and the validation loss is reported using the best one. 
The same seed is used for all experiments. Codebase available at \url{https://github.com/lucaslingle/mu_transformer}.

\subsubsection{Configuration}

We use the following default configuration for the experiments, deviating only when specifically mentioned.
The depth is fixed at $L = 24$ and we consider model widths $M \in \{128, 512, 2048\}$, yielding three model sizes ranging from 4.7M to 1.2B non-embedding parameters.\footnote{We consider using $4\times$ deeper models briefly in Appendix \ref{deeper_models}.}
The head width is fixed at $D = 128$, the number of heads is $H = M/D$, and MLP hidden width is $F = 4M$. The models by default use nonparametric RMSNorm \citep{ZhangSennrich2019RMSNorm, Groeneveld2024}, linear projections without biases \citep{Raffel2020T5}, RoPE on the queries and keys \citep{Su2021Rotary}, and ReLU for the MLP activation \citep{Vaswani2017Transformer, Raffel2020T5}. 

By default, we use $2^{18}$ tokens per batch, float32 parameters, and bfloat16 activations and gradients\footnote{During evaluation, output logits are computed in float32.}. 
The optimizer is AdamW \citep{Loshchilov2017AdamW} with $\beta_{1} = 0.9, \beta_{2} = 0.98, \epsilon = 10^{-9}$, with default weight decay $0.0$ and gradient clip $1.0$. Models train for 125K steps total, with 10K steps of learning rate warmup followed by linear decay to zero. 

\subsection{Ablation Experiments}

In these experiments, we sweep the base learning rate $\alpha \in \{ 2^{-2j} : j \in \mathbb{N}, 1 \leq j \leq 5 \}$ for each model size and experiment setting, and we report all results. This allows us to investigate the impact of each experimental condition on model quality and learning rate transferability. We focus on learning rate transfer because it is the main hyperparameter of interest for large transformer models.

\subsubsection{$\mu$P Baseline}

In this group of experiments, we establish $\mu$P baselines to facilitate comparison in later ablations. In addition, we verify that $\mu$-Transfer works reliably even with mixed-precision training. Following \citep{Rae2022Gopher, Hoffmann2022Chinchilla}, we utilize the Google Brain floating point format, bfloat16, for the activations and gradients. This format is supported by Google TPUs and recent Nvidia GPUs, and previously used by \citep{Dey2023CerebrasGPT}. Recent work \citep{Ahmadian2023CohereQuantization} suggests it may reduce emergent outliers versus float16 format \citep{Dettmers2022Emergent, Elhage2023Privileged}. 

As shown in Table \ref{mu-table}, the learning rates transfer reliably across model sizes via $\mu$P. Despite each model being over an order of magnitude larger than the previous, the smallest model's optimal base learning rate $\alpha$ directly predicts the optimum in the sweeps for the larger model sizes. 

\subsubsection{Biases}

It is not clear if bias vectors in linear layers are beneficial for model quality, and several prior works omit them \citep{Raffel2020T5, Shazeer2020GLU, Chowdhery2023Palm}. Here we test their benefit and impact on learning rate transferability, keeping all other details identical to the baseline $\mu$P settings. Following \citet{Yang2022TP5}, the learning rate for the biases under $\mu$P is given as $\Theta(1)$, so we use $\alpha$ for the exact learning rate. 

As shown in Table \ref{mu-table}, the learning rates do transfer across model sizes via $\mu$P. However, including biases in the linear layers does not reliably improve quality over the $\mu$P baseline models. 

\subsubsection{Parametric RMSNorm}

It is not clear if scale vectors (`gains') in RMSNorm \citep{ZhangSennrich2019RMSNorm} are beneficial for model quality and many frameworks offer the option to omit them. This ablation tests their benefit and impact on learning rate transferability under $\mu$P. 
We also test a variant where the trainable gain vector is replaced with a trainable scalar multiplier, similar to \citet{Elhage2023Privileged} and in concurrent work by \citet{Everett2024}. 

As shown in Table \ref{mu-table}, optimal learning rates for these models {do not reliably transfer} when using $\alpha = \Theta(1)$ learning rate scaling for the gains as dictated by \citet{Yang2022TP5}. 

Intriguingly however, when comparing to the baseline $\mu$P results, we find trainable gains harm the quality of the largest $\mu$P models when the base learning rate $\alpha$ is optimal. This suggests trainable gains might be simply omitted in $\mu$P models without loss of quality at scale. Moreover, in Section \ref{sp_experiment} we find $\mu$P transformers without trainable gains \emph{outperform} standard transformers using them. 

\subsubsection{Query Initialization}

The usual $\mu$P initialization for query projections $\mathbf{W}^{AQ}$ is i.i.d. Gaussian with variance $\Theta(1/M)$, but zero-initialization was an alternative discussed by \citet{Yang2022TP5} so we investigate it here. 

As shown in Table \ref{mu-table}, the learning rates transfer across model sizes when using $\mu$P with zero-initialized query projections. There also appears to be a slight yet consistent improvement in loss. 

\subsubsection{Standard Attention Scale}

The usual attention scale $\tau^{-1} = 1/\sqrt{D}$ was proposed by \citet{Vaswani2017Transformer} and generally used since. However, $\mu$P proposes $\tau^{-1} = \Theta(1/D)$, and we use $\tau^{-1} = 1/D$. Notably in our experiments we scale the model width $M$ and keep the head width $D$ fixed across model sizes, so the attention scale would presumably not actually matter for purposes of transfer, since any difference between $1/\sqrt{D}$ and $1/D$ can be treated as a constant multiplier in the model architecture. 

Nonetheless, as shown in Table \ref{mu-table}, the usual $1/\sqrt{D}$ attention scale appears quite suboptimal, yielding worse performance than the corresponding $\mu$P baselines and preventing transfer of the learning rate optimum from the proxy model with $M = 128$. In Appendix \ref{extra-heads-appdx}, we similarly obtain a lack of transfer with the usual $1 / \sqrt{D}$ attention scale for $D = 32$. The lack of transfer is resolved for both settings $D \in \{32, 128\}$ when using float32 precision and averaging over three seeds, but not when averaging over three randoms seeds alone (Table \ref{multiseed-table}). This suggests that choices of constant multiplier may impact learning rate transfer from very narrow widths $M$ when training in low-precision formats. 

\subsubsection{Standard Unembedding Initialization}

The $\mu$P initialization for unembedding matrix $\mathbf{W}^{U}$ uses a i.i.d. Gaussian random variables with variance $\Theta(1/M^{2})$, while the so-called standard parameterization (STP) uses variance $1/M$ \citep{Yang2022TP5}. We thus ablate the impact of using the standard initialization on performance and transfer. 

As shown in Table \ref{mu-table}, despite using this initialization for the unembeddings, the learning rate optimum empirically transfers across models. A similar result was also reported by \citet{Everett2024}. 

\subsubsection{Cosine Schedule}

The linear learning rate schedule used as our default configuration is one of several possible choices \citep{Touvron2023Llama2, Chen2023Lion, DeepseekAI2024DeepSeekLLM, Hu2024MiniCPM}. In this ablation, we instead use a cosine schedule decaying to zero. 

As shown in Table \ref{mu-table}, both schedules perform similarly and the learning rates transfer across model sizes, showing that $\mu$P is empirically compatible with a cosine schedule as well as a linear schedule. 

\subsubsection{Weight Decay}
\label{wd-experiment-section}

In libraries such as Pytorch and Optax, the AdamW optimizer \citep{Loshchilov2017AdamW} essentially takes a schedule $\{s_{t}\}_{t=1}^{T}$ and a hyperparameter $\lambda$, and computes the decay coefficient as $\eta \lambda s_{t}$ rather than using $\lambda s_{t}$, where $\eta$ is the learning rate for a given parameter tensor. The latter `independent weight decay' is advocated by \citep{Wortsman2023SmallScale, Littwin2023TP4b}.

We study independent weight decay in Table \ref{iwd-table}. We observe the optimal $(\alpha, \lambda)$ appears to transfer, but for one fixed choice of decay, $\lambda = 10^{-3}$, the optimal $\alpha$ alone did not. This was possibly due to training instabilities in the larger models not observed in the proxy model, and suggests that optimizing $(\alpha, \lambda)$ may be important for independent weight decay. 

We now consider the version of weight decay in these libraries. As shown in Table \ref{mu-table}, when using this version of weight decay, $\alpha$ also did not transfer from the smallest model to the larger ones. On the other hand, these experiments used the typical setting at $\lambda = 0.1$, yet the optimal $\alpha$ was not measurably changed between the larger models. The large-model optimum for $\alpha$ also appears to match the optimum for the baseline $\mu$P models with no weight decay. 
Based on these observations, a heuristic strategy for using this version of weight decay would be to omit it when training the proxy model, and apply it to the target model only. 

\subsubsection{Embedding Normalization}

We consider using normalized embeddings following \citep{Peng2023RWKV, GemmaTeam2024}, but using nonparametric RMSNorm \citep{ZhangSennrich2019RMSNorm, Groeneveld2024} because parametric RMSNorm interfered with learning rate transfer in our previous ablations. We do not change the learning rate from the setting in Table 2 nor adjust the initialization. 

As shown in Table \ref{mu-table}, the optimal learning rate transfers across models using $\mu$P; however, the improvement in model quality over the baseline is negligible. We briefly also investigated using lower initialization variances, but found this harmed stability with the width-constant embedding learning rate of $\mu$P Adam, perhaps due to similar impacts on training dynamics. 

\subsubsection{Multiplicative Nonlinearities}

Multiplicative nonlinearities such as SwiGLU \citep{Shazeer2020GLU} and Squared ReLU \citep{So2021Primer} are increasingly used in the MLP blocks to improve transformer quality \citep{Touvron2023Llama, Touvron2023Llama2, Elsen2023Persimmon, Peng2023RWKV, Parmar2024Nemotron4}. In this experiment, we investigate both of the aforementioned nonlinearities, which are notably `superlinear' and thus may create outliers that interfere with $\mu$-Transfer, as discussed by \citet{Yang2022TP5}. 
For SwiGLU, we use $F = 5M$, so the MLP has $7.5M^{2}$ parameters. 

As shown in Table \ref{mu-table}, the SwiGLU and Squared ReLU nonlinearities both allow $\mu$-transfer of the learning rate across model sizes. These outcomes contrast with parametric RMSNorm experiments, since learning rate transfer occurs despite the presence of the multiplicative interactions. 

\subsection{Multi-Query Attention}

Multi-Query Attention \citep{Shazeer2019MQA} and its grouped generalization \citep{Ainslie2023GQA} are increasingly used in transformer LLMs \citep{Chowdhery2023Palm, Touvron2023Llama2, Almazrouei2023Falcon, GeminiTeam2023Gemini1, Jiang2024Mixtral}. 
These techniques aim to improve the inference speed of transformers by sharing keys/values across multiple heads. This ablation investigates the impact on $\mu$-Transfer. 
Similar to \citet{Shazeer2019MQA}, we approximately correct for the parameter increase by setting $F = 5M$. 

As shown in Table \ref{mu-table}, multi-query attention is empirically compatible $\mu$-Transfer. 

\subsection{Lion Optimizer}

We empirically investigate if the Lion optimizer \citep{Chen2023Lion, Chen2024Lyapunov} is compatible with $\mu$-Transfer. This optimizer is at least twice as memory-efficient as the Adam optimizer, and was reported to yield models of similar quality, including transformers \citep{Chen2023Lion}. A notable property of this optimizer is that its updates are constrained to $\{-1, +1\}$ per coordinate, yielding a coordinate size of $\Theta(1)$ per step. Thus, a $\Theta(1/M)$ scaling for hidden weight learning rates, similar to $\mu$P Adam, may be appropriate \citep{Yang2022TP5}. 
 
As shown in Table \ref{mu-table}, the Lion optimizer did not admit transfer of the base learning rate from the smallest model size. The scaling rules do appear to preserve the optimum $\alpha$ between larger models. When averaging over three random seeds for the proxy model (Table \ref{multiseed-table}), transfer is restored. 

\subsubsection{4x Smaller Batch}

An important question is if $\mu$-Transfer requires a minimum batch size to work \citep{Dey2023CerebrasGPT}. In this section, we consider scaling the batch size down by $4\times$, while keeping the number of training tokens the same. 
For this ablation, we apply the scaling rule from \citet{You2020LAMB, Malladi2022SDE}, so that each learning rate formula in Table 2 is scaled by $0.5 \times$. 

As shown in Table \ref{mu-table}, the $4\times$ smaller batch size admits transfer of the learning rate via $\mu$P. Nonetheless, further research on the permissible batch sizes for $\mu$-Transfer seems advisable. 

\subsubsection{4x Larger Batch}

Large-batch training can reduce wall time, but may also have a considerable influence on the training dynamics \citep{McCandlish2018LargeBatch, You2020LAMB}. In this section, we consider scaling up the batch size by $4\times$ while keeping the number of training tokens the same. For this ablation, we again apply the scaling rule from \citet{You2020LAMB, Malladi2022SDE}, so that each learning rate formula in Table 2 is scaled by $2 \times$. 

As shown in Table \ref{mu-table}, the $4\times$ larger batch size admits transfer of the learning rate via $\mu$P. 

\subsection{Standard Parameterization Experiment}
\label{sp_experiment}

In this experiment, we consider the observation from \citet{Yang2022TP5} that $\mu$P models outperform models using the standard parameterization (STP). In particular, the STP transformer studied here uses trainable biases in linear layers, parametric RMSNorm layers, attention scale $1 / \sqrt{D}$, and unembedding initialization variance $1 / M$. Adam is used for optimization, and the learning rate $\eta$ for all parameter tensors is equal. This learning rate for each model size is swept over in the table below. All other hyperparameters are identical to the $\mu$P baseline. 

\begin{table}[h]
\centering
\caption{Validation losses for STP models.} 
\label{sp-table}
\begin{tabular}{rccccccc}
\toprule
\multicolumn{1}{c}{Width ($M$)} &\multicolumn{5}{c}{LR ($\eta$)} \\
\cmidrule{2-6}
& $2^{-10}$ & $2^{-8}$ & $2^{-6}$ & $2^{-4}$ & $2^{-2}$ & \\
\midrule
128 & 3.841 & 3.757 & \bf \textcolor{black}{3.706} & 3.879 & 4.030 &  \\
512 & 3.013 & \bf \textcolor{black}{2.967} & 2.987 & 3.383 & 7.403 & \\
2048 & \bf \textcolor{black}{2.738} & 2.902 & 7.247 & 7.477 & 7.314 & \\
\bottomrule
\end{tabular}
\end{table}

As shown in Tables \ref{sp-table} and \ref{sp-highres-table}, for all model sizes, the optimal loss in the sweep is higher than the optimal loss for the baseline $\mu$P models shown in Table \ref{mu-table}. This result may suggest that using a width-constant learning rate for the embeddings, nonparametric RMSNorm, and the $\mu$P attention scale help improve optimization. 

\subsection{Scaling Up}
\label{large_experiment}

In this experiment, we combine the architectural choices that improved performance while allowing learning rate transfer, and we investigate if $\mu$-Transfer continues to work at a larger scale. 

We use depth $L = 12$ and consider widths $M \in \{128, 512, 2048, 8192 \}$, yielding models with approximately 2M, 40M, 600M, and 10B non-embedding parameters, respectively. 
We use zero-initialized queries \citep{Yang2022TP5} and Squared ReLU nonlinearity \citep{So2021Primer}. 
We use $2^{21}$ tokens per batch, training for 90K steps. We use AdamW \citep{Loshchilov2017AdamW} with 
$\beta_{1} = 0.9, \beta_{2} = 0.95, \epsilon = {10}^{-8}$, decoupled weight decay $\lambda = 0.1$, and gradient clip $1.0$. 

\begin{table}[h]
\centering
\caption{Validation losses for our largest-scale experiment.} 
\label{largescale-table}
\begin{tabular}{rrccc}
\toprule
\multicolumn{1}{c}{Params} & \multicolumn{1}{c}{Width ($M$)} &\multicolumn{3}{c}{Base LR ($\alpha$)}  \\
\cmidrule{3-5}
& & $2^{-8}$ & $2^{-6}$ & $2^{-4}$ \\
\midrule
2M & 128 & 3.791 & \bf \textcolor{black}{3.766} & 3.814  \\
40M & 512 & 3.016 & \bf \textcolor{black}{2.983} & 3.004 \\
600M & 2048 & 2.513 & \bf \textcolor{black}{2.459} & 2.466  \\
10B & 8192 & 2.238 & \bf \textcolor{black}{2.167} & 2.169 \\
\bottomrule
\end{tabular}
\end{table}

As shown in Table \ref{largescale-table}, the optimal learning rate in the sweep transfers from a model $4000\times$ smaller. Thus, $\mu$-Transfer continues to predict the optimal learning rate at this model scale and LR granularity. This result may also suggest that emergent outlier features do not significantly interfere with learning rate transfer at this scale, given that prior work reports outlier features appearing at approximately 7B parameters in some open-source models \citep{Dettmers2022Emergent}. 

As shown in Table \ref{largescale-highres-table}, when increasing the granularity of the LR sweep to $2\times$, there is a slight drift in the optimum $\alpha$ as a function of model width. A similar drift also occurred in prior experiments using the same form of weight decay (Table \ref{mu-table}). Disabling weight decay in proxies, an informal heuristic from Section \ref{wd-experiment-section}, led to better transfer of $\alpha$ to the largest model with weight decay (Table \ref{largescale-highres-table}). 

Future works may wish to consider using independent weight decay \citep{Littwin2023TP4b, Wortsman2023SmallScale} instead of our heuristic. The sweep done in Table \ref{iwd-table} seems to indicate that the $(\alpha, \lambda)$ optimum would transfer, but verifying this for $M = 8192$ would make our final experiment several times more expensive. In addition, switching to the Lion optimizer \citep{Chen2023Lion} or applying one of the Adam-only methods from \citet{Littwin2023TP4b, Everett2024} may also help reduce drift in the $\alpha$ optimum. 

\section{Related Works}

The $\mu$-Parameterization ($\mu$P) is proposed in the Tensor Programs series \citep{Yang2019TP1, Yang2020TP2, Yang2021TP3, Yang2021TP4, Yang2022TP5, Littwin2023TP4b, Yang2023TP6} and independently derived for various cases in related work \citep{Bordelon2022DMFT, Yang2024SpectralCondition}. 

The empirical observation of hyperparameter transfer via $\mu$P was first reported by \citet{Yang2022TP5}. Their largest target model had 6.7B parameters, trained in float32 precision versus mixed precision baselines, and used a different position encoding and learning rate schedule than the baselines. It was shown to outperform 6.7B and 13B parameter baselines. However despite the impressive performance, their experiments did not specifically show $\mu$P preserved the hyperparameter optimum of the target model, which may be difficult due to the number of hyperparameters where transfer was applied. In this work, we scale to 10B parameters, use mixed precision with optimized baselines, and evaluate learning rate transfer only. 

Some recent works have adopted $\mu$P for hyperparameter tuning \citep{Dey2023CerebrasGPT, Dey2023BTLM3B8K, Hu2024MiniCPM, Hu2024Predicting, XAI2024Grok1}, but do not show that the hyperparameter optimum is preserved under $\mu$P in larger-scale settings (e.g., around 10B parameters and 200B tokens). For example, \citet{Dey2023CerebrasGPT} trained a suite of models, but only used $\mu$P for up to 2.7B parameters, while the largest had 13B parameters. This left open the question of whether $\mu$P accurately transfers the learning rate optimum to very large target models. 

Several recent works suggest to apply $\mu$P alongside specific ideas such as depthwise scaling \citep{Jelassi2023Depthwise, Bordelon2023Depthwise, Yang2023TP6, Chen2024Principled} or loss prediction \citep{Yao2023LossPrediction, Fan2024NanoLM}. However, similar to the other works, these papers do not focus on evaluating $\mu$P for width-only transferability of hyperparameter optima. 

An alternative to $\mu$P for hyperparameter tuning can be found in \citet{DeepseekAI2024DeepSeekLLM}, where scaling laws for optimal batch size and learning rate are proposed in terms of compute budget. 
Additional related work considers transformer training instabilities via small proxy models \citep{Wortsman2023SmallScale}, and find methods that reduce the loss sensitivity to the learning rate. Other works \citep{Bernstein2023AGD, Baydin2017Hypergradient, Chandra2019Hypergradient, Defazio2023DAdaptation, Mishchenko2023Prodigy} propose to tune the learning rate optimizers during training; \citet{Defazio2024ScheduleFree} proposes a `schedule-free' optimizer requiring only a constant learning rate. From a theoretical standpoint, \citet{Noci2023Shaped} proposes a transformer parameterization with a well-defined limit over depth, width, and time; \citet{Noci2024SuperConsistency} describes a connection between learning rate transfer and the stability of the top eigenvalue of the training loss Hessian across model sizes. Other related work studies the role of batch size \citep{McCandlish2018LargeBatch, Shallue2019}, learning rate schedule \citep{Zhang2019NQM, Geiping2022Cramming, Defazio2024Schedule}, optimizer \citep{Nado2021LargeBatchOptimizers, Kaddour2023NoTrainNoGain}, and training steps \citep{Bjorck2024Horizon}. 

Concurrently to this work, \citet{Everett2024} studies the scaling behavior of learning rate optima for several initialization schemes, including a version of $\mu$P using embeddings multiplied by $\sqrt{M}$ and a version of the standard parameterization, and considers the SGD, Adam, and Adafactor optimizers. Their paper used RMSNorm with scalar parameters \citep{Elhage2023Privileged}, which we studied in an ablation experiment. Similar to another of our experiments, that paper finds the unembedding projection can be initialized with the standard $\mathcal{N}(0, 1/M)$ without observing a negative impact on performance. In addition, the authors consider mitigations for Adam learning rate drift not studied here, and train $2\times$ wider models than our widest. However, our paper considers a different range of experiments: our smallest models are over $20\times$ smaller, all models are $1.5$-$3\times$ deeper, and our largest models train on $28\times$ more tokens (190B vs 6B) and are approximately compute-optimal under the Chinchilla scaling law \citep{Hoffmann2022Chinchilla}. 

\section{Conclusion}

This paper studies the reliability of $\mu$-Transfer of the learning rate as proposed by \citet{Yang2022TP5} and applied to transformer language models. In our extensive ablations, transfer worked as promised in most cases, though not when using trainable RMSNorm parameters or the standard attention scale. The most effective workarounds for these cases were to use nonparametric RMSNorm and $1/D$ attention scale, both of which improved model performance and enabled learning rate transfer. We also found the standard implementation of weight decay can prevent transfer, and found omitting decay from the proxy or switching to independent weight decay \citep{Littwin2023TP4b, Wortsman2023SmallScale} can improve transferability. 

The $\mu$P strategy used in this work also outperformed the `standard parameterization' when the latter included parametric RMSNorm and standard attention scale. Lastly, we found $\mu$-Transfer from a 2M parameter model approximately predicted the optimal learning rate from a sweep at the scale of 10B parameters, though we observed a slight drift in the $\alpha$ optimum across the $4000\times$ scale increase. 

\begin{ack}
The author sincerely thanks the EleutherAI community and NeurIPS reviewers for helpful remarks. Project supported with Cloud TPUs from Google's TPU Research Cloud. 
\end{ack}

\bibliography{neurips_2024}
\clearpage

\appendix
\section{Appendix}

\subsection{Ablations}
\begin{table}[H]
\centering
\caption{Validation losses for $\mu$-transfer ablations. Best loss for each model size is shown in bold.} 
\label{mu-table}
\begin{tabular}{lrcccccc}
\toprule
\multicolumn{1}{c}{Ablation} & \multicolumn{1}{c}{Width ($M$)} &\multicolumn{5}{c}{Base LR ($\alpha$)} &\multicolumn{1}{c}{Transfer} \\
\cmidrule{3-7}
& & $2^{-10}$ & $2^{-8}$ & $2^{-6}$ & $2^{-4}$ & $2^{-2}$ & \\
\midrule
 & 128 & 3.846 & 3.743 & \bf \textcolor{black}{3.695} & 3.884 & 4.143 &  \\
Our Baseline $\mu$P & 512 & 3.114 & 2.993 & \bf \textcolor{black}{2.953} & 3.221 & 3.506 & \cmark \\
 & 2048 & 2.711 & 2.553 & \bf \textcolor{black}{2.511} & 2.563 & 3.244 &  \\
\midrule
 & 128 & 3.838 & 3.735 & \bf \textcolor{black}{3.705} & 3.911 & 4.269 &  \\
 Projection Biases & 512 & 3.108 & 2.986 & \bf \textcolor{black}{2.947} & 2.970 & 3.557 & \cmark \\
 & 2048 & 2.710 & 2.552 & \bf \textcolor{black}{2.529} & 2.672 & 3.418 &  \\
  \midrule
 & 128 & 3.842 & 3.744 & 3.689 & \bf \textcolor{black}{3.670} & 3.681 &  \\
Vector RMSNorm Params & 512 & 3.101 & 2.992 & 2.951 & \bf \textcolor{black}{2.950} & 3.412 & \xmark \\
 & 2048 & 2.692 & \bf \textcolor{black}{2.553} & 2.609 & 2.605 & 3.169 &  \\
\midrule
 & 128 & 3.843 & 3.749 & 3.692 & \bf \textcolor{black}{3.670} & 4.471 &  \\
Scalar RMSNorm Params & 512 & 3.106 & 3.000 & 2.961 & \bf \textcolor{black}{2.959} & 3.515 &  \xmark  \\
 & 2048 & 2.704 & 2.570 & \bf \textcolor{black}{2.525} & 2.542 & 3.334 &  \\
\midrule
 & 128 & 3.836 & 3.743 & \bf \textcolor{black}{3.694} & 3.877 & 4.167 &  \\
Zero Query Init & 512 & 3.115 & 2.992 & \bf \textcolor{black}{2.949} & 3.135 & 3.532 & \cmark \\
 & 2048 & 2.711 & 2.553 & \bf \textcolor{black}{2.510} & 2.551 & 3.272 &  \\
  \midrule
 & 128 & 3.836 & \bf \textcolor{black}{3.758} & 3.905 & 4.140 & 4.597 &  \\
Standard Attention Scale & 512 & 3.104 & 2.993 & \bf \textcolor{black}{2.962} & 3.449 & 4.184 &  \xmark  \\
 & 2048 & 2.706 & 2.555 & \bf \textcolor{black}{2.525} & 3.306 & 7.280 &  \\
 \midrule
 & 128 & 3.861 & 3.765 & \bf \textcolor{black}{3.699} & 3.896 & 4.161 &  \\
Standard Unembedding Init & 512 & 3.119 & 2.990 & \bf \textcolor{black}{2.951} & 3.265 & 3.582 & \cmark  \\
 & 2048 & 2.716 & 2.554 & \bf \textcolor{black}{2.509} & 2.564 & 7.471 &  \\
 \midrule
 & 128 & 3.846 & 3.743 & \bf \textcolor{black}{3.695} & 3.906 & 4.143 &  \\
Cosine Schedule & 512 & 3.114 & 2.995 & \bf \textcolor{black}{2.955} & 3.225 & 3.506 & \cmark \\
 & 2048 & 2.712 & 2.558 & \bf \textcolor{black}{2.518} & 2.572 & 3.244 &  \\
  \midrule
 & 128 & 3.760 & \bf \textcolor{black}{3.679} & 3.694 & 3.741 & 4.011 &  \\
Weight Decay & 512 & 3.057 & 2.963 & \bf \textcolor{black}{2.957} & 3.139 & 3.373 & \xmark \\
 & 2048 & 2.686 & 2.535 & \bf \textcolor{black}{2.502} & 3.123 & 6.594 &  \\
  \midrule
& 128 & 3.834 & 3.743 & \bf \textcolor{black}{3.693} & 4.012 & 4.120 &  \\
Embedding Normalization & 512 & 3.115 & 2.993 & \bf \textcolor{black}{2.954} & 3.028 & 3.506 &  \cmark  \\
 & 2048 & 2.710 & 2.553 & \bf \textcolor{black}{2.512} & 2.564 & 7.316 &  \\
  \midrule
 & 128 & 3.800 & 3.740 & \bf \textcolor{black}{3.715} & 4.090 & 7.024 &  \\
SwiGLU Nonlinearity & 512 & 3.070 & 2.975 & \bf \textcolor{black}{2.953} & 3.175 & 6.863 &  \cmark  \\
 & 2048 & 2.677 & 2.536 & \bf \textcolor{black}{2.505} & 2.553 & 4.571 &  \\
 \midrule
 & 128 & 3.808 & 3.735 & \bf \textcolor{black}{3.686} & 3.999 & 4.484 &  \\
Squared ReLU Nonlinearity & 512 & 3.071 & 2.964 & \bf \textcolor{black}{2.929} & 3.184 & 7.299 &  \cmark  \\
 & 2048 & 2.666 & 2.516 & \bf \textcolor{black}{2.482} & 2.532 & 3.259 &  \\
  \midrule
& 128 & \bf \textcolor{black}{3.708} & 3.736 & 4.057 & 4.344 & 10.380 &  \\
Lion Optimizer & 512 & 2.952 & \bf \textcolor{black}{2.947} & 3.416 & 3.961 & 10.285 &  \xmark  \\
& 2048 & 2.519 & \bf \textcolor{black}{2.511} & 3.151 & 10.377 & 10.377 &  \\
   \midrule
 & 128 & 3.811 & 3.708 & \bf \textcolor{black}{3.667} & 3.881 & 4.121 &  \\
Multi-Query Attention & 512 & 3.101 & 2.979 & \bf \textcolor{black}{2.940} & 3.187 & 3.518 &  \cmark  \\
 & 2048 & 2.715 & 2.564 & \bf \textcolor{black}{2.521} & 2.546 & 3.257 &  \\
\midrule
& 128 & 3.855 & 3.774 & \bf \textcolor{black}{3.736} & 3.945 & 4.104 &  \\
4x Smaller Batch & 512 & 3.120 & 3.011 & \bf \textcolor{black}{2.977} & 3.024 & 3.521 & \cmark  \\
& 2048 & 2.714 & 2.568 & \bf \textcolor{black}{2.527} & 2.549 & 3.223 &  \\
\midrule
& 128 & 3.844 & 3.735 & \bf \textcolor{black}{3.697} & 3.716 & 10.380 &  \\
4x Larger Batch & 512 & 3.141 & 2.990 & \bf \textcolor{black}{2.965} & 3.305 & 10.373 & \cmark  \\
& 2048 & 2.745 & 2.556 & \bf \textcolor{black}{2.541} & 2.697 & 7.197 &  \\
\bottomrule
\end{tabular}
\end{table}

\subsection{Increased Attention Heads}
\label{extra-heads-appdx}

In this section, we investigate the effect of using extra attention heads per transformer layer, using a narrower head width $D = 32$ with $H = M/D$ to keep the parameter count equal. We investigate the effect of extra heads for the baseline configuration and the ablations where transfer between the proxy model and the next largest model size failed. 

\begin{table}[h]
\centering
\caption{Validation losses for selected $\mu$-transfer ablations with extra heads; $L=24$.}
\label{extra-heads-table}
\begin{tabular}{lrcccccc}
\toprule
\multicolumn{1}{c}{Group} & \multicolumn{1}{c}{Width ($M$)} &\multicolumn{5}{c}{Base LR ($\alpha$)} &\multicolumn{1}{c}{Transfer} \\
\cmidrule{3-7}
& & $2^{-10}$ & $2^{-8}$ & $2^{-6}$ & $2^{-4}$ & $2^{-2}$ & \\
\midrule
 Our Baseline $\mu$P & 128 & 3.813 & 3.720 & \bf \textcolor{black}{3.678} & 3.997 & 7.519 &  \\
 & 512 & 3.106 & 2.985 & \bf \textcolor{black}{2.951} & 3.005 & 6.962 & \cmark \\
 \midrule
 STP Attention Scale &128 & 3.798 & \bf \textcolor{black}{3.716} & 3.865 & 4.105 & 7.273 &  \\
 & 512 & 3.097 & 2.986 & \bf \textcolor{black}{2.958} & 3.451 & 7.158 &  \xmark  \\
 \midrule
 Decoupled Weight Decay &128 & 3.722 & \bf \textcolor{black}{3.650} & 3.660 & 3.708 & 3.919 &  \\
 & 512 & 3.049 & 2.951 & \bf \textcolor{black}{2.945} & 3.218 & 3.414 & \xmark \\
\midrule
Lion Optimizer & 128 & 3.676 & \bf \textcolor{black}{3.666} & 4.092 & 10.380 & 10.380 &  \\
& 512 & 2.943 & \bf \textcolor{black}{2.942} & 3.421 & 10.373 & 10.373&  \cmark  \\
\bottomrule
\end{tabular}
\end{table}

As shown in Table \ref{extra-heads-table}, using $D = 32$ did not qualitatively alter our transfer results, with the exception of the Lion optimizer experiment, where transfer now appears to work. 

\subsection{Standard Parameterization, Increased Sweep Granularity}
\label{sp-highres-appdx}

In this section, we validate the findings presented in the main text for the STP transformer model, utilizing a $2\times$ granularity learning rate sweep and expanding the range of learning rates in both directions until the loss stops decreasing. 

\begin{table}[h]
\centering
\caption{Validation losses for STP models; $L=24$.} 
\label{sp-highres-table}
\begin{tabular}{ccccccccccccc}
\toprule
\multicolumn{1}{c}{Width ($M$)} &\multicolumn{10}{c}{LR ($\eta$)} \\
\cmidrule{2-11}
& $2^{-13}$ & $2^{-12}$ & $2^{-11}$ & $2^{-10}$ & $2^{-9}$ & $2^{-8}$ & $2^{-7}$ & $2^{-6}$ & $2^{-5}$ & $2^{-4}$ & \\
\midrule
128 & 4.187 & 4.036 & 3.917 & 3.841 & 3.792 & 3.757 & 3.726 & \bf \textcolor{black}{3.706} & 3.724 & 3.879 &  \\
512 & 3.264 & 3.127 & 3.051 & 3.013 & 2.988 & 2.967 & \bf \textcolor{black}{2.961} & 2.987 & 3.185 & 3.383 & \\
2048 & 2.623 & \bf \textcolor{black}{2.574} & 2.605 & 2.738 & 2.739 & 2.902 & 2.853 & 7.247 & 7.142 & 7.477 & \\
\bottomrule
\end{tabular}
\end{table}

Similar to the lower sweep resolution, we do not observe transfer, and moreover the pattern for the optimal $\alpha$ does not appear to be linear on a log-log scale. In addition, for $M = 2048$ with optimal $\alpha$, the loss is quite a bit worse than the Standard Attention Scale or Parametric RMSNorm ablations alone, and is also worse than the Baseline $\mu$P model (Table \ref{mu-table}). 

\subsection{Scaling Up, Increased Sweep Granularity}
\label{largescale-highres-appdx}

\begin{table}[h]
\centering
\caption{Validation losses for our largest experiment; $L=12$.}
\label{largescale-highres-table}
\begin{tabular}{rrrccccc}
\toprule
\multicolumn{1}{c}{Params} & \multicolumn{1}{c}{Width ($M$)} & \multicolumn{1}{c}{Weight Decay ($\lambda$)} &\multicolumn{5}{c}{Base LR ($\alpha$)}  \\
\cmidrule{4-8}
& & & $2^{-8}$ & $2^{-7}$ & $2^{-6}$ & $2^{-5}$ & $2^{-4}$ \\
\midrule
2M & 128 & 0.0 & 3.870 & 3.843 & 3.819 & \bf \textcolor{black}{3.803} &  3.805  \\
\midrule
2M & 128 & 0.1 & 3.791 & 3.768 & \bf \textcolor{black}{3.766} & 3.773 & 3.814  \\
40M & 512 & 0.1 & 3.016 & 2.996 & \bf \textcolor{black}{2.983} & 2.985 & 3.004 \\
600M & 2048 & 0.1 & 2.513 & 2.477 & 2.459 & \bf \textcolor{black}{2.456} & 2.466  \\
10B & 8192 & 0.1 & 2.238 & 2.190 & 2.167 & \bf \textcolor{black}{2.161} & 2.169 \\
\bottomrule
\end{tabular}
\end{table}

\subsection{Absolute Scaling Rules}
\label{abs-baseline-appdx}

In this section, we investigate using absolute scaling rules, dividing the hidden weights' learning rate by their fan-in width instead of by a width-expansion ratio relative to a proxy model. 

While these learning rate scaling rules are equivalent up to architecture-dependent constants, one practical difference is the learning rates for hidden weights are no longer equal, since each is now computed as $\eta = \alpha / \texttt{fan\_in}$. 
\begin{table}[h]
\centering
\caption{Validation losses for absolute scaling $\mu$-transfer ablations; $L=24$.}
\label{abs-baseline-table}
\begin{tabular}{lrccccc}
\toprule
\multicolumn{1}{c}{Ablation} & \multicolumn{1}{c}{Width ($M$)} &\multicolumn{4}{c}{Base LR ($\alpha$)} &\multicolumn{1}{c}{Transfer} \\
\cmidrule{3-6}
& & $2^{-1}$ & $2^{1}$ & $2^{3}$ & $2^{5}$ & \\
\midrule
Abs. $\mu$P & 128 & 3.687 &  \bf \textcolor{black}{3.664} & 3.677 & 4.191 & \\
 & 512 & 2.986 & \bf \textcolor{black}{2.950} & 2.966 & 3.579 & \cmark \\
  & 2048 & 2.588 & \bf \textcolor{black}{2.538} & 3.029 & 3.776 & \\
\bottomrule
\end{tabular}
\end{table}

\subsection{Ablations using Multiple Trials}
\label{multiseed-appdx}

To assess whether the results obtained in Table \ref{mu-table} are representative of each experimental condition, we run the smallest experiments again using three random seeds and display the results in Table \ref{multiseed-table}. 

As shown in Table \ref{multiseed-table}, for the ablations using Parametric RMSNorm, Standard Attention Scale, and Weight Decay, the learning rate optimum using the averaged loss for $M=128$ does not transfer to the optima for $M > 128$ in Table \ref{mu-table}. This lack of transfer is consistent with the single-seed results for $M=128$ for these settings, as given in Table \ref{mu-table}.

For STP Attention Scale with fixed $D \in \{32, 128\}$, float32 precision resolves the lack of transfer. 
For Lion with $D = 128$ transfer is resolved using three random seeds alone. 
For Lion with $D = 32$, the optimum is unclear due to the high standard error of the validation loss for some $\alpha$. 

\begin{table}[h]
\centering
\caption{Validation losses with $L=24$ and $M=128$. We report the mean $\pm$ stddev over 3 seeds.} 
\label{multiseed-table}
\begin{tabular}{lrrrrr}
\toprule
\multicolumn{1}{c}{Group} &\multicolumn{5}{c}{Base LR ($\alpha$)} \\
\cmidrule{2-6}
& $2^{-10}$ & $2^{-8}$ & $2^{-6}$ & $2^{-4}$ & $2^{-2}$ \\
\midrule
Our Baseline $\mu$P, D=128 & 3.840 $\pm$ 0.006 & 3.744 $\pm$ 0.003 & \bf \textcolor{black}{3.695} $\pm$ \bf \textcolor{black}{0.002} & 3.908 $\pm$ 0.013 & 4.195 $\pm$ 0.045 \\
Our Baseline $\mu$P, D=32 & 3.813 $\pm$ 0.001 & 3.713 $\pm$ 0.004 & \bf \textcolor{black}{3.695} $\pm$ \bf \textcolor{black}{0.037} & 3.901 $\pm$ 0.002 & 7.589 $\pm$ 0.380 \\
Projection Biases & 3.838 $\pm$ 0.002 & 3.744 $\pm$ 0.007 & \bf \textcolor{black}{3.702} $\pm$ \bf \textcolor{black}{0.002} & 3.806 $\pm$ 0.077 & 5.317 $\pm$ 1.717 \\
Vector RMSNorm Params & 3.833 $\pm$ 0.007 & 3.743 $\pm$ 0.002 & 3.689 $\pm$ 0.003 & \bf \textcolor{black}{3.677} $\pm$ \bf \textcolor{black}{0.007} & 3.845 $\pm$ 0.086 \\
Scalar RMSNorm Params & 3.835 $\pm$ 0.007 & 3.750 $\pm$ 0.001 & 3.694 $\pm$ 0.003 & \bf \textcolor{black}{3.679} $\pm$ \bf \textcolor{black}{0.004} & 4.218 $\pm$ 0.166 \\
Zero Query Init & 3.837 $\pm$ 0.003 & 3.743 $\pm$ 0.002 & \bf \textcolor{black}{3.692} $\pm$ \bf \textcolor{black}{0.001} & 3.909 $\pm$ 0.030 & 4.358 $\pm$ 0.350 \\
STP Attn, D=128 & 3.832 $\pm$ 0.003 & \bf \textcolor{black}{3.756} $\pm$ \bf \textcolor{black}{0.002} & 3.893 $\pm$ 0.032 & 4.147 $\pm$ 0.022 & 8.990 $\pm$ 1.303 \\
STP Attn, D=32 & 3.797 $\pm$ 0.004 & \bf \textcolor{black}{3.717} $\pm$ \bf \textcolor{black}{0.003} & 3.829 $\pm$ 0.016 & 4.108 $\pm$ 0.013 & 8.338 $\pm$ 1.778 \\
STP Attn, D=128, fp32 & 3.831 $\pm$ 0.002 & 3.755 $\pm$ 0.001 & \bf \textcolor{black}{3.719} $\pm$ \bf \textcolor{black}{0.004} & 4.106 $\pm$ 0.029 & 4.918 $\pm$ 0.286 \\
STP Attn, D=32, fp32 & 3.795 $\pm$ 0.005 & 3.716 $\pm$ 0.004 & \bf \textcolor{black}{3.685} $\pm$ \bf \textcolor{black}{0.003} & 4.094 $\pm$ 0.003 & 5.675 $\pm$ 0.042 \\
STP Unembedding Init & 3.853 $\pm$ 0.006 & 3.755 $\pm$ 0.008 & \bf \textcolor{black}{3.699} $\pm$ \bf \textcolor{black}{0.002} & 3.940 $\pm$ 0.057 & 4.154 $\pm$ 0.016 \\
Cosine Schedule & 3.840 $\pm$ 0.005 & 3.745 $\pm$ 0.003 & \bf \textcolor{black}{3.696} $\pm$ \bf \textcolor{black}{0.002} & 3.916 $\pm$ 0.014 & 4.195 $\pm$ 0.045 \\
Weight Decay & 3.750 $\pm$ 0.008 & \bf \textcolor{black}{3.681} $\pm$ \bf \textcolor{black}{0.005} & 3.691 $\pm$ 0.001 & 3.752 $\pm$ 0.003 & 4.003 $\pm$ 0.009 \\
Emb Normalization & 3.836 $\pm$ 0.002 & 3.744 $\pm$ 0.003 & \bf \textcolor{black}{3.698} $\pm$ \bf \textcolor{black}{0.004} & 4.049 $\pm$ 0.038 & 4.178 $\pm$ 0.065 \\ 
SwiGLU  & 3.801 $\pm$ 0.003 & 3.741 $\pm$ 0.002 & \bf \textcolor{black}{3.722} $\pm$ \bf \textcolor{black}{0.008} & 4.064 $\pm$ 0.047 & 6.566 $\pm$ 0.506 \\
Squared ReLU & 3.806 $\pm$ 0.005 & 3.732 $\pm$ 0.002 & \bf \textcolor{black}{3.695} $\pm$ \bf \textcolor{black}{0.008} & 3.976 $\pm$ 0.056 & 5.177 $\pm$ 0.921 \\
Lion, D=128 & 3.706 $\pm$ 0.002 & \bf \textcolor{black}{3.693} $\pm$ \bf \textcolor{black}{0.009} & 4.074 $\pm$ 0.025 & 6.546 $\pm$ 3.318 & 10.382 $\pm$ 0.006 \\
Lion, D=32 & \bf \textcolor{black}{3.674} $\pm$ \bf \textcolor{black}{0.002} & 3.688 $\pm$ 0.049 & 4.084 $\pm$ 0.002 & 10.382 $\pm$ 0.006 & 10.382 $\pm$ 0.006 \\
Multi-Query Attn & 3.801 $\pm$ 0.006 & 3.713 $\pm$ 0.006 & \bf \textcolor{black}{3.673} $\pm$ \bf \textcolor{black}{0.005} & 3.947 $\pm$ 0.058 & 4.135 $\pm$ 0.028 \\
4x Larger Batch & 3.846 $\pm$ 0.004 & 3.741 $\pm$ 0.007 & \bf \textcolor{black}{3.694} $\pm$ \bf \textcolor{black}{0.008} & 4.901 $\pm$ 2.033 & 10.382 $\pm$ 0.006 \\
4x Smaller Batch & 3.857 $\pm$ 0.005 & 3.774 $\pm$ 0.002 & \bf \textcolor{black}{3.735} $\pm$ \bf \textcolor{black}{0.001} & 3.965 $\pm$ 0.012 & 4.126 $\pm$ 0.005 \\
\bottomrule
\end{tabular}
\end{table}

\subsection{Independent Weight Decay}

\begin{table}[H]
\centering
\caption{Validation losses for independent weight decay; $L=24$.} 
\label{iwd-table}
\begin{tabular}{rrcccc}
\toprule
\multicolumn{1}{c}{WD ($\lambda$)} & \multicolumn{1}{c}{Width ($M$)} &\multicolumn{4}{c}{Base LR ($\alpha$)} \\
\cmidrule{3-6}
& & $2^{-10}$ & $2^{-8}$ & $2^{-6}$ & $2^{-4}$ \\
\midrule
 & 128 & 3.864 & 3.728 & 3.673 & \bf \textcolor{black}{3.666} \\
$10^{-3}$ & 512 & 3.270 & 3.089 & \bf \textcolor{black}{2.997} & 3.312 \\
 & 2048 & 2.948 & 2.730 & \bf \textcolor{black}{2.646} & 2.647  \\
\midrule
 & 128 & 3.759 & 3.684 & \bf \textcolor{black}{3.661} & 3.936 \\
$10^{-4}$ & 512 & 3.031 & 2.959 & \bf \textcolor{black}{2.927} & 3.311 \\
 & 2048 & 2.626 & 2.535 & \bf \textcolor{black}{2.498} & 2.528  \\
 \midrule
  & 128 & 3.835 & 3.734 & \bf \textcolor{black}{3.689} & 3.877 \\
  $10^{-5}$ & 512 & 3.094 & 2.984 & \bf \textcolor{black}{2.946} & 3.211 \\
   & 2048 & 2.682 & 2.543 & \bf \textcolor{black}{2.503} & 2.542 \\
\bottomrule
\end{tabular}
\end{table}

\subsection{Increased Layers}
\label{deeper_models}

\begin{table}[H]
\centering
\caption{Validation losses for deep models, with zero-initialized residual blocks; $L=96$.} 
\label{deep-table}
\begin{tabular}{rrcccc}
\toprule
\multicolumn{1}{c}{Width ($M$)} &\multicolumn{4}{c}{Base LR ($\alpha$)} \\
\cmidrule{2-5}
& $2^{-11}$ & $2^{-9}$ & $2^{-7}$ & $2^{-5}$ \\
\midrule
128 & 3.565 & 3.495 & \bf \textcolor{black}{3.456} & 3.640 \\
512 & 2.894 & 2.789 & \bf \textcolor{black}{2.753} & 2.848 \\
\bottomrule
\end{tabular}
\end{table}

\clearpage

\end{document}